\documentclass[letterpaper, conference]{IEEEtran}  
\usepackage{graphicx}
\usepackage{url}
\usepackage[disable]{todonotes}
\usepackage{pgfplots} 
\usepackage{pifont}
\usepackage{gensymb}
\newcommand{\figname}{Figure\,}%
\newcommand{\andothers}{et al.\,}%
\usepackage[official]{eurosym}

\IEEEoverridecommandlockouts                              

\pgfplotsset{width=.3\linewidth,compat=1.9}

\title{\LARGE \bf
Scratchy: A Lightweight Modular Autonomous Robot for Robotic Competitions
}

\author{Raphael Memmesheimer, Isabelle Kuhlmann, Mark Mints, Patrik Schmidt,\\ Christian Korbach, Ida Germann and Dietrich Paulus
\thanks{University of Koblenz-Landau,
        56070 Koblenz, Germany
        {\tt\small raphael@uni-koblenz.de}}%
}

\begin{document}

\maketitle

\IEEEpubid{978-1-7281-3558-8/19/ \$31.00 \copyright\ 2019 IEEE}


\begin{abstract}

We present \textit{Scratchy}---a modular, lightweight robot built for low budget competition attendances. Its base is mainly built with standard 4040 aluminium profiles and the robot is driven by four mecanum wheels on brushless DC motors.  In combination with a laser range finder we use estimated odometry -- which is calculated by encoders -- for creating maps using a particle filter. A RGB-D camera is utilized for object detection and pose estimation. Additionally, there is the option to use a 6-DOF arm to grip objects from an estimated pose or generally for manipulation tasks. The robot can be assembled in less than one hour and fits into two pieces of hand luggage or one bigger suitcase. Therefore, it provides a huge advantage for student teams that participate in robot competitions like the European Robotics League or RoboCup. Thus, this keeps the funding required for participation, which is often a big hurdle for student teams to overcome, low. The software and additional hardware descriptions are available under: \url{https://github.com/homer-robotics/scratchy}.
\end{abstract}
\IEEEpubidadjcol

\section{INTRODUCTION}
Robot competitions have the potential of bringing robotic systems from controlled lab environments into realistic everyday environments. As Behnke et al. \cite{behnke2006robot} argue, they even are an ideal platform for 
benchmarking robotic systems.

\begin{figure}[t] \centering$
  \vspace{0.06in}
  \begin{array}{cc}
      \includegraphics[height=.75\linewidth]{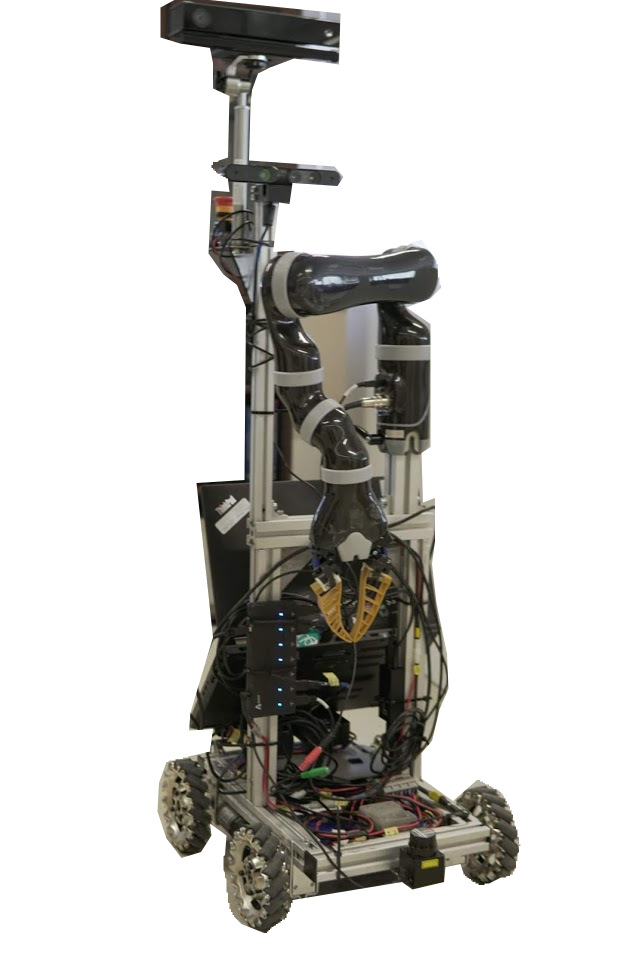} &
      \includegraphics[height=.75\linewidth]{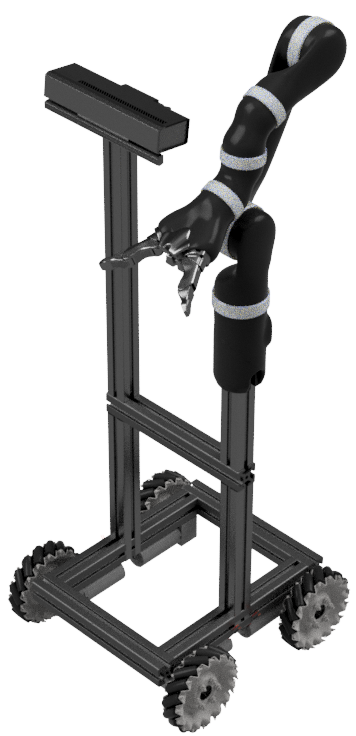} \\
      (a) & (b) 
  \end{array}$
  \caption{Fully assembled \textit{Scratchy} $(a)$, as used during the local tournament of the European Robotics League Consumer Service Robots and a rendered view $(b)$}
  \label{fig:robot}
\end{figure}

However, they evoke obstacles that are often of financial nature. International competitions like the RoboCup come along with high costs for robot transportation. Oftentimes this leads to
teams initially planning to compete but having to resign from their participation due to budget issues. Therefore, teams are often excluded or limited in their amount of competitions to take part in. We try to tackle
these problems by proposing the robot \textit{Scratchy}  (see \figname \ref{fig:robot}), which is modular and can be transported in carry-on luggage only. Moreover, we prove it to be functional during a participation in the European Robotics League (ERL). 


Robotic competitions are a motivating environment for students. 
Hence, an ideal platform would be low on cost, easy and cheap to transport (even over high distances) and 
highly modular in order to allow for different sensor and actuator setups. Furthermore, the robot should be robust and should not be prone to scratches. These factors tend to keep students from experimenting. Broken parts like motors should be replaceable by students
or researchers and should be affordable. These requirements exclude almost all research platforms available. 
\textit{Scratchy} is meant as a modular platform that will be extended over time. First, the robot can be used to navigate using a laser range finder and motor encoders. Afterwards, a RGB-D camera can be attached to localize objects. Once objects can be localized, it can be extended with an arm in order to manipulate objects. On the other hand, if one wants to focus more on a human-robot interaction level, it can be extended with a more appealing case: a monitor for showing expressions, a camera for detection and recognition of humans and -- in order to communicate -- a microphone that can be attached.
The base of \textit{Scratchy} is defined as a mobile autonomous platform that is able to build maps and navigate within an apartment. This basic functionality is provided by accompanied packages. Advanced functionalities can be build on top of these packages.
\IEEEpubidadjcol
The contribution of this paper is as follows:
\begin{itemize}
    \item We propose a lightweight, modular robotic setup named \textit{Scratchy}, targeted at low cost participation in robot competitions.
    \item Furthermore, we publish packages to use the robot for autonomous mobile service robotic operations. This includes configuration files and libraries for odometry generation, mapping and navigation, manipulation and text to speech interaction.
\end{itemize}
This paper is structured as follows: First, we present related work in the robotic competition field and 
introduce robotic platforms used in both, robot competitions and research labs. We then introduce \textit{Scratchy}, present the proposed hardware and software solutions and give insight into our design decisions. Next, an evaluation of \textit{Scratchy} during the European Robotics League for Consumer Service Robots is shown. The paper ends with a discussion of the mentioned topics.

\section{RELATED WORK}

For robot competitions you find a variety of robotic setups that are used by the 
individual teams. In research labs, there are some common robot platforms 
\cite{cousins2010ros, fitzgerald2013developing, pages2016tiago, bischoff2011kuka} 
established already. Most of them are quite expensive, and, for example the Kuka YouBot \cite{bischoff2011kuka},
the Rethink Baxter Robot \cite{fitzgerald2013developing} (as of 01/2019) and the PR2 \cite{fitzgerald2013developing} are already unsupported or only vaguely supported by their respective companies. 
In 2017, RoboCup@Home has introduced standard platforms in separate leagues, namely the Human Support Robot (HSR) \cite{yamaguchi2015hsr} and the Social Standard Platform with the aim  of making teams focus on the tasks rather than the hardware. Limited processing power and network dependency have caused a lot of issues since the introduction. A standardized platform is also existing for the RoboCup soccer scenario \cite{gouaillier2009mechatronic} which led to the development of approaches for the platforms limited in computing power.
More recently, Paull \andothers \cite{paull2017duckietown} presented Duckietown, an open research platform for miniature autonomous driving tasks. Duckietown consists of Duckiebots, which are the autonomous robots themselves, driving around in Duckietown. A whole ecosystem was created, not only consisting of the hardware and software, but also providing lecture materials, assignments and guidelines to interact with Duckiebots and to develop new features. The main focus in development with Duckietown is that all components are easy to set up and easy to use, as well as being cheap. Duckiebots support different levels of education starting from tasks for undergraduate students, like line following and simple marker detection, but also challenging objectives, like dealing with collision avoidance and other complex behaviours while taking limited computational resources into account. 

Piperidis \cite{piperidis2007low} proposed a modular low cost platform for research and development. The platform is differential and built on minimalist low budget parts.
Our proposed design is approximately double the size in order to manipulate objects like domestic furniture and interact with humans. Further, we follow a mecanum base platform to be more flexible during manipulation tasks where adjusting the robot using a differential design usually requires a sequence of linear and rotary motions.





\section{HARDWARE DESIGN}


\begin{figure}[t] \centering
  \vspace{0.06in}
  \includegraphics[width=\linewidth]{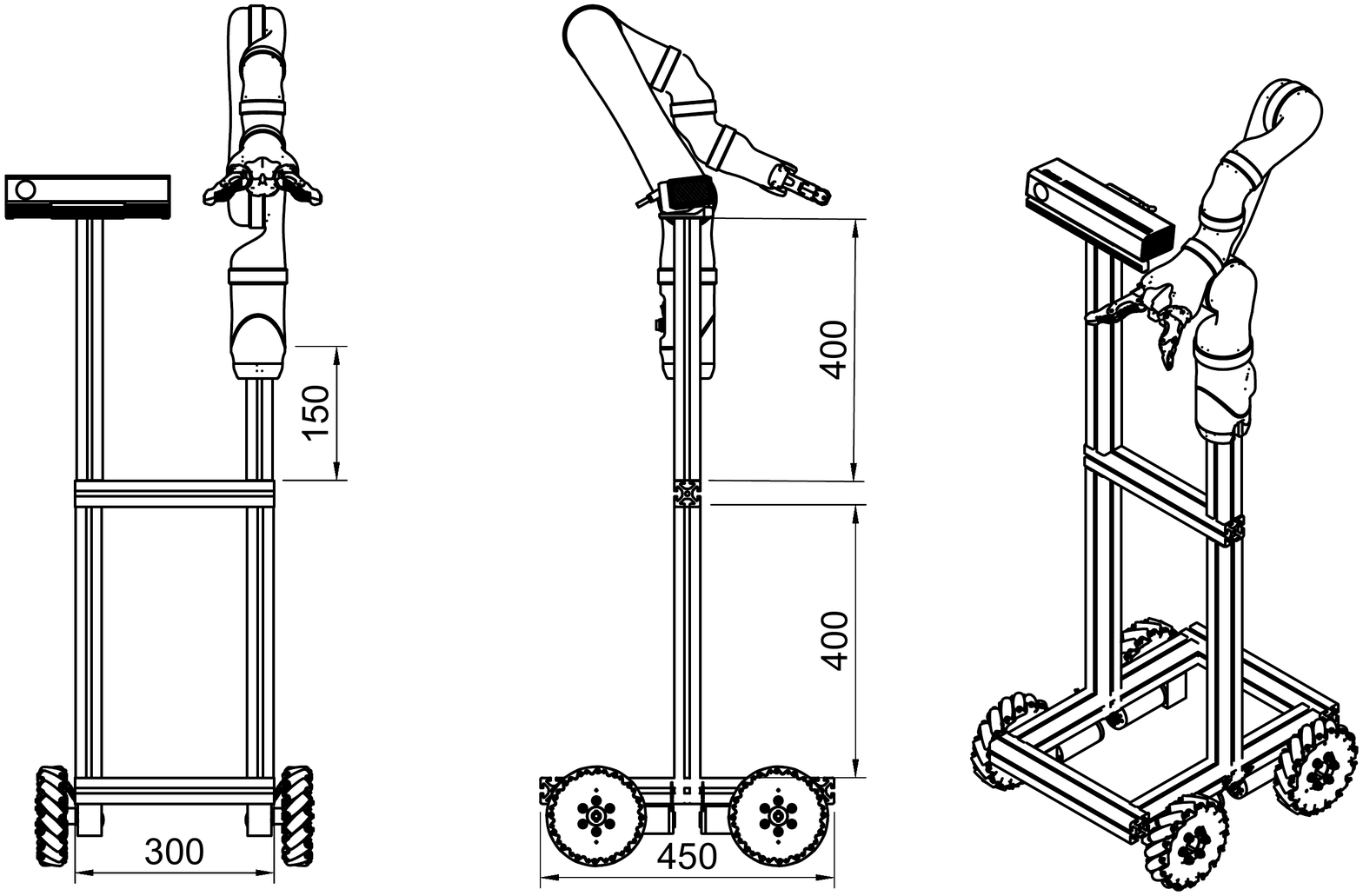} 

  \caption{Technical drawings of \textit{Scratchy}. \todo{add}}
  \label{fig:technical}
\end{figure}

\begin{table}[]
\caption{Data}
\centering
\begin{tabular}{|l|l|}
\hline
\textbf{Data}  & \textbf{Value}   \\ \hline
Length  & 45\,cm   \\ \hline
Width  & 35\,cm   \\ \hline
Height  & 140\,cm   \\ \hline
Weight (without arm) & 16\,kg   \\ \hline
Weight (with arm)  & 21\,kg   \\ \hline
\end{tabular}
\label{tab:datasheet}
\end{table}
In this section we describe the hardware design suggested by us and inspect the used
components, which are interchangeable. Technical drawings are given in \figname \ref{fig:technical}. Most parts are designed based on aluminium profiles with standard screw connectors. Sensors and manipulators can often be found in research labs. Measurements and weights are given in Table \ref{tab:datasheet}. The lower part of the robot consists of a rectangular frame that mounts four motors with attached encoders. Mecanum wheels provide the ability to drive omnidirectional. The motors are controlled by a standard Arduino Mega micro-controller board. Two  motor driver shields (Pololu Dual VNH5019) -- one shield operating two motors -- are connected to the Arduino and serve as controllers for the motors. \textit{Scratchy} is powered by two DJI TB47 drone batteries serving 24V with self-protection and can be carried in hand luggage, too---avoiding any restrictions due to flight safety of most airlines. An power converter yields 12V for operating additional devices like the Microsoft Kinect 2. We advice to connect the RGB-D camera directly via USB to the laptop which operates the robot, due to the high demand on bandwidth. The Hokuyo URG-04LX-UG01 in combination with the wheel encoders adds the ability to create occupancy grid-maps for navigation and to avoid obstacles. 
 However, it is important to provide some sort of housing to protect the laser scanner from damage during operation, especially when experimenting with different navigation algorithms and clearance levels. The mounting position and orientation of the  manipulator is  crucial for the operation range. Lower mounting positions will increase the overall stability of the platform and scan positions for the RGB-D cameras are unlikely to be blocked by any parts of the manipulator. The Kinova Mico manipulator was mounted on 4040 aluminium profiles. The manipulation range can be influenced by experimenting with other mounting positions easily.

One of the main goals was to create a small scaled robotic base with the aim to minimize shipping costs. All aluminium profiles have a maximal length of 40\,cm. All parts together weigh a little less than 16\,kg excluding the arm. Since, apart from the accumulators, there are no parts which require special safety precautions, transport in hand luggage states no problem.
The final assembling took around 40 minutes by 2 persons while the disassembling took only 11 minutes by 3 persons. Just three different tools were needed for the assembling. A component list is given in Table \ref{tab:components}. Further drawings can be found on the project page.

\begin{table}[]
\caption{Components}
\centering
\begin{tabular}{|l|l|l|r|}
\hline
\textbf{Description}  & \textbf{Reference}            & \textbf{Level} & \textbf{Cost\ \euro{}}  \\ \hline
LRF              & Hokuyo URG-04LX-UG01           & Autonomy  & 1.000 \\ \hline 
Notebook              & Lenovo Thinkpad P50           & Autonomy  & 1.500 \\ \hline 
USB Hub  &  CSL – USB 3.0 Hub                    & Autonomy  & 30  \\ \hline
Controller              & Arduino Mega              & Autonomy & 20 \\ \hline 
Motor Controller              & 2 x Pololu Dual VNH5019             & Autonomy & 80 \\ \hline
Battery   &  TB41                   & Autonomy & 150  \\ \hline
Wheels & 4 x 152mm Mecanum  & Autonomy & 150 \\ \hline
Motors              & KAG M42x40/I+SNR17             & Autonomy & 400 \\ \hline 
RGB-D Camera   &  Microsoft Kinect 2                   & Perception & 150  \\ \hline

Microphone      &  Rode VidMic GO          & HRI & 50 \\ \hline
Speaker       & Xiaomi Mi BT 2           & HRI & 20\\ \hline
Manipulator       & Kinova Mico 6\,DOF          &  Manipulation & 24.000\\ \hline
\end{tabular}
\label{tab:components}
\end{table}

\section{SOFTWARE DESIGN}

The proposed hardware is functionally integrated into a software design that is
easy to use and to extend. We provide integration into the widely used Robot Operating System (ROS) \cite{quigley2009ros}. Autonomous operation including a mapping and navigation component, basic human-robot interaction by a text to speech integration is provided. This yields a great base for a
competition and research platform to build on top. The source code and documentation are provided on \url{https://github.com/homer-robotics/scratchy}.

\todo{include basic node graph}

\subsection{Mecanum Odometry}

\newcommand{\front}{\uparrow}
\newcommand{\rear}{\downarrow}
\newcommand{\links}{\leftarrow}
\newcommand{\rechts}{\rightarrow}
\newcommand{\distance}{d}
\newcommand{\dfl}{d_{\front,\links}}
\newcommand{\dfr}{d_{\front,\rechts}}
\newcommand{\drl}{d_{\rear,\links}}
\newcommand{\drr}{d_{\rear,\rechts}}
\newcommand{\wheeldiff}{\Delta w}

The four encoders attached to the motor axis yield the wheel revolutions in ticks with 100\,Hz.
Given the ticks per wheel and location $t_{\front,\links}$, $t_{\front,\rechts}$, $t_{\rear,\links}$ , $t_{\rear,\rechts}$, we 
can compute the corresponding distances of the revolution by measuring the ticks for a fixed distance $d$ resulting in a distance per tick $d_t$ and multiply them with the corresponding amount of ticks like:
\begin{equation}
  \dfl = t_{\front,\links} \times d_t,
\end{equation}
where $\front$ and $\links$ denote the distance for the front left wheel. The other corresponding 
distances are computed likewise.


To compute the odometry we further need the averaged distance between the wheels $\beta$, where $\wheeldiff$ is defined as the distance between the wheels
\begin{equation}
\beta = \frac{1}{4\wheeldiff}
\end{equation}

from which we can compute the change in orientation

\begin{equation}
    \Delta \theta = (-\dfl + \dfr - \drl + \drr) * \beta.
\end{equation}
Similar we compute the relative ticks for $x,y$ with:
\begin{equation}
\Delta s_x = \frac{\dfl  + \dfr + \drl  + \drr}{4}
\end{equation}
and
\begin{equation}
\Delta s_y = \frac{-\dfl  + \dfr + \drl - \drr}{4}.
\end{equation}
This allows us to compute the relative pose based on the current encoder readings as
\begin{equation}
  \Delta x = \Delta s_x * cos(\theta + \frac{\Delta \theta}{2}) + \Delta s_y * cos(\theta + \frac{\Delta \theta + \pi}{2}),
\end{equation}
 
\begin{equation}
  \Delta y = \Delta  s_y * sin(\theta + \frac{\Delta \theta}{2}) + \Delta s_y * cos(\theta + \frac{\Delta \theta + \pi}{2}).
\end{equation}
  
We can then compute the overall robot pose in 2D by incrementally summing up all local information yielding the global 2D pose ($x$ , $y$ for the position and $\theta$ for the orientation): 

\begin{equation}
    x = x + \Delta x
\end{equation}
\begin{equation}
    y = y + \Delta y
\end{equation}
\begin{equation}
   \theta = \theta + \Delta \theta.
\end{equation}
The odometry serves as input for the drift step in the mapping process.

\subsection{Mapping}

\begin{figure}
    \centering
    \includegraphics[width=.7\linewidth]{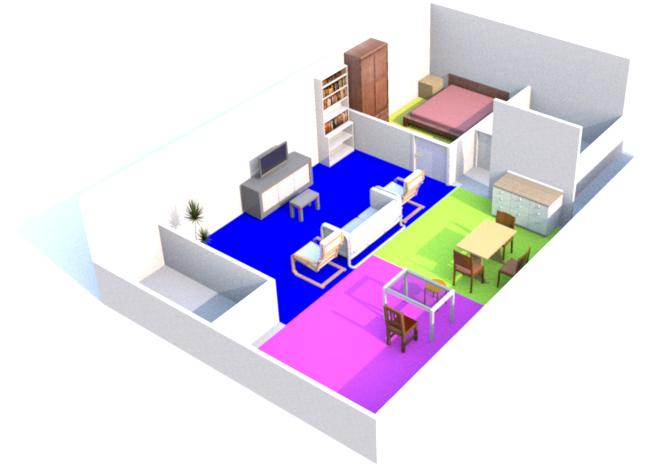}
    \caption{A 3D representation of the ISRoboNet@Home test bed.}
    \label{fig:3d_map}
\end{figure}

\begin{figure}
    \centering
    \includegraphics[width=.7\linewidth]{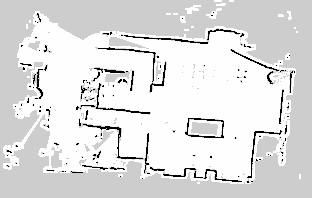}
    \caption{The generated map of the ISRoboNet@Home test bed is shown.}
    \label{fig:slam_map}
\end{figure}

Our mapping approach is based on the approach by Pellenz \andothers \cite{pellenz2008mapping}.
For that reason, a particle filter -- which uses 1000 particles and contains resamples, drifts and measures -- is used as a SLAM approach to create a $35 m^2$ map with a resolution of $0.05 m$ per cell. 
To update the particle weights when resampling, the robot has to move at least $10mm$ or turn $5\degree$. After executing the motion model in the drift, we update the particle weights by the calculated odometry. Next, the particles are spread by assuming a translational error of
$2\%$ and a rotational error of $3\%$. These values are estimated by moving the 
robot a fixed distance and turning it by a fixed angle. By measuring  the error between the real movements and the estimated odometry we can determine the error values. 
The average pose of the top 5\% weighted particles is taken as the estimated pose
which defines how the current measurement is used for extending the map by its relative
pose difference to the previous pose. The resulting map of the testbed as used during the competition is shown in Figure \ref{fig:slam_map}.

\subsection{Navigation}

In real-life situations, a static grid map is not sufficient
for navigating through an everyday environment, as due to the movement 
of persons and other dynamic obstacles, an occupancy map that only changes slowly in time
does not provide sufficient information \cite{seib2015robocup}.

Thus, our navigation system, which is based on Zelinsky's path transform  \cite{Zelinsky1988RNW},
always merges the current laser
range scan as a frontier into the occupancy map. 
If an object blocks the path for a given interval, the path is re-calculated.
This approach allows for the robot to efficiently navigate and to avoid obstacles in highly dynamic environments \cite{seib2015robocup,Memmesheimer2016R2H}.

\subsection{Text To Speech}

For text to speech we interface with Marry TTS \cite{DBLP:conf/lrec/PammiCS10}. A HTTP server runs on the accompanied notebook, and requested text to speech is streamed directly to the audio output. We use the \textit{cmu-slt} voice model. 

\section{EVALUATION}

\begin{figure*}[t] \centering$
  \vspace{0.06in}
  \begin{array}{ccc}
      \begin{tikzpicture}
\begin{axis}[
    title={TBM1 Results},
    xlabel={Round},
    ylabel={Count},
    xmin=1, xmax=5,
    ymin=0, ymax=8,
    xtick={1,2,3,4,5,6,7},
    ytick={0,1,2,3,4,5,6,7,8},
    legend pos=north west,
    ymajorgrids=true,
    grid style=dashed,
]
\addplot[color=green, mark=square] coordinates {(1,3)(2,5)(3,4)(4,6)(5,4)};\addlegendentry{A}
\addplot[color=red, mark=square] coordinates {(1,0)(2,0)(3,1)(4,0)(5,1)};\addlegendentry{P}
\end{axis}
\label{fig:tbm1_results}
\end{tikzpicture} &
\begin{tikzpicture}
\begin{axis}[
    title={TBM2 Results},
    xlabel={Round},
    ylabel={Count},
    xmin=1, xmax=4,
    ymin=0, ymax=9,
    xtick={1,2,3,4,5,6,7},
    ytick={0,1,2,3,4,5,6,7,8,9},
    legend pos=north west,
    ymajorgrids=true,
    grid style=dashed,
]
\addplot[color=green, mark=square] coordinates {(1,2)(2,5)(3,6)(4,6)};\addlegendentry{A}
\addplot[color=red, mark=square] coordinates {(1,1)(2,1)(3,0)(4,0)};\addlegendentry{P}
\end{axis}
\end{tikzpicture}
&
\begin{tikzpicture}
\begin{axis}[
    title={TBM3 Results},
    xlabel={Round},
    ylabel={Count},
    xmin=1, xmax=6,
    ymin=0, ymax=8,
    xtick={1,2,3,4,5,6,7},
    ytick={0,1,2,3,4,5,6,7,8},
    legend pos=north west,
    ymajorgrids=true,
    grid style=dashed,
]
\addplot[color=green, mark=square] coordinates {(1,3)(2,3)(3,4)(4,4)(5,4)(6,4)};\addlegendentry{A}
\addplot[color=red, mark=square] coordinates {(1,2)(2,0)(3,1)(4,1)(5,1)(6,2)};\addlegendentry{P}
\end{axis}
\end{tikzpicture}\\
      \begin{tikzpicture}
\begin{axis}[
    title={TBM4 Results},
    xlabel={Round},
    ylabel={Count},
    xmin=1, xmax=4,
    ymin=0, ymax=9,
    xtick={1,2,3,4,5,6,7},
    ytick={0,1,2,3,4,5,6,7,8,9},
    legend pos=north west,
    ymajorgrids=true,
    grid style=dashed,
]
\addplot[color=green, mark=square] coordinates {(1,6)(2,2)(3,4)(4,6)};\addlegendentry{A}
\addplot[color=red, mark=square] coordinates {(1,0)(2,1)(3,0)(4,0)};\addlegendentry{P}
\end{axis}
\label{fig:tbm1_results}
\end{tikzpicture} 
&
\begin{tikzpicture}
\begin{axis}[
    title={FBM2 Results},
    xlabel={Round},
    ylabel={Error},
    xmin=1, xmax=3,
    ymin=0, ymax=0.5,
    xtick={1,2,3},
    ytick={0.1,0.2,0.3,0.4,0.5},
    legend pos=north west,
    ymajorgrids=true,
    grid style=dashed,
]
\addplot[color=green, mark=square] coordinates {(1,0.31)(2,0.11)(3,0.12)};\addlegendentry{Translation}
\addplot[color=red, mark=square] coordinates {(1,0.27)(2,0.0031)(3,0.09)};\addlegendentry{Rotation}
\end{axis}
\end{tikzpicture}
  \end{array}$
  \caption{Competition results of all task and functionality benchmarks. Achievements (A) are plotted in green while penalties (P) are plotted in red.}
  \label{fig:results}
\end{figure*}
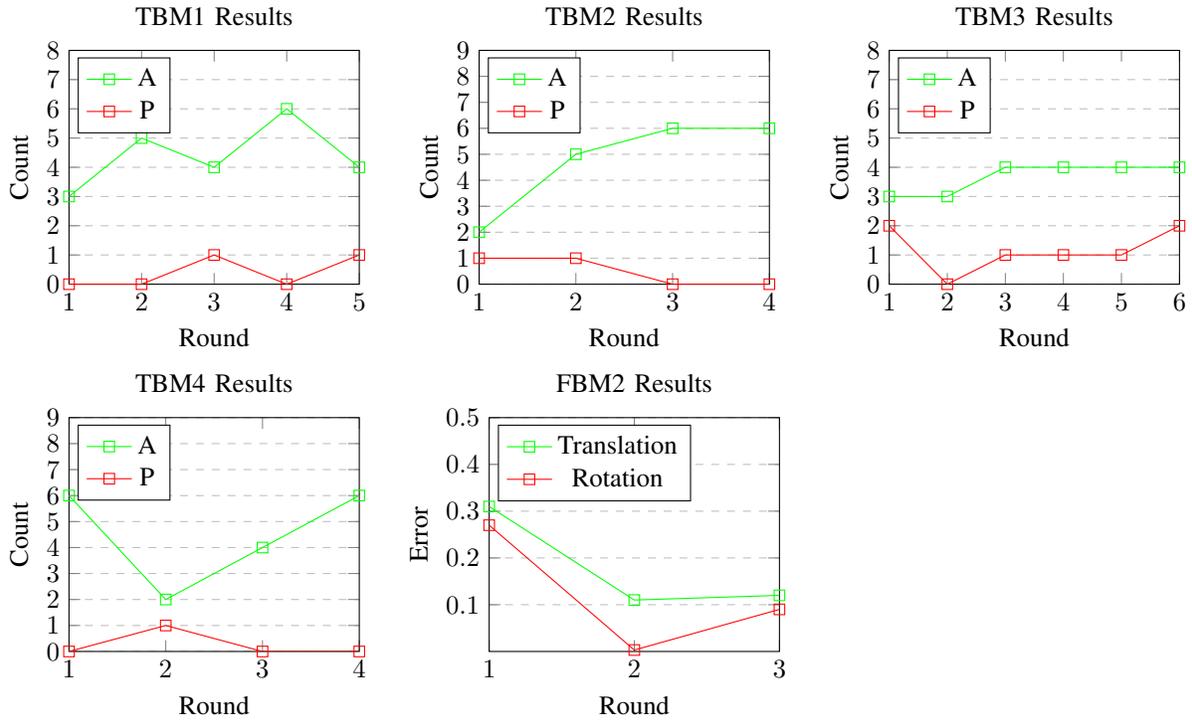

We evaluated our robot design at a participation of the European Robotics League
Consumer local tournament 2019 in Lisbon, Portugal, at the ISRoboNet@Home test bed located in the premises of the Institute for Systems and Robotics of Instituto Superior T\'{e}cnico, University of Lisbon. 

\subsection{Test bed} 

The local tournament took place in ISRoboNet@Home test bed, which we will briefly introduce now. A more detailed description can be found on a dedicated page\footnote{\url{http://welcome.isr.tecnico.ulisboa.pt/isrobonet/}}. For benchmarking the robots,
the test bed is equipped with 12 OptiTrack PRIME13 motion capturing cameras. The ground truth positions of reflective marker sets are attached to the robots to estimate poses. The testbed mimics a real apartment (a rendering is given in \figname \ref{fig:3d_map}). It is divided into different rooms which are furnished realistically. A map created by \textit{Scratchy} is given in \figname \ref{fig:slam_map}.

\subsection{Tasks}

The ERL for Consumer Robots \cite{lima2016rockin} is divided into the following functionality benchmarks:
\begin{itemize}
    \item FBM1: Object Perception
    \item FBM2: Navigation
\end{itemize}

and the following task benchmarks:
\begin{itemize}
    \item TBM1: Getting to Know My Home
    \item TBM2: Welcoming Visitors
    \item TBM3: Catering for Granny Annie's Comfort
    \item TBM4: Visit My Home.
\end{itemize}
Regulations are described in a rulebook \cite{erl_rulebook}, which differs for each league.
Functionality benchmarks focus on lower level capabilities which are reused and extended to practical real world scenarios in the task benchmarks.

The first task benchmark \textit{Getting to Know my Home} (TBM1) is focused on finding one's way in a new dynamic environment and to register changes like moved furniture or objects. These were noted in a semantic map. To enable the robot to orientate itself as quickly as possible in an apartment, in addition to the metric map there is a semantic knowledge base that describes the relationships between rooms, furniture, and objects.
In \textit{Welcoming Visitors} (TBM2) the robot has to interact with a door bell of a smart home, which initiates the start of this task benchmark. A variety of visitors are visiting the apartment and therefor, the robot has to recognize the postman and the doctor by face recognition. Unknown persons have to be handled by asking who they are and right after the identification, the visitors are guided into a specific room of the apartment --- depending on their business. After they have finished their
job, the robot has to accompany them back to the exit. So, the main focus of this task is on navigation and person recognition. 
\textit{Catering for Grannie Annie} (TBM3) benchmarks the speech recognition and understanding abilities of a robot. Therefor, a speech command consisting of three sub commands --- which could contain manipulation tasks like bringing an object from a furniture piece, accompany a person or searching a person or an object in a room are given and should be executed by the robot.
\textit{Visit My Home} (TBM4) is about navigation in a changing apartment. The robot is asked to go to three different locations, but the path to go there is blocked and an alternative path has to be found. Further a path having only one way is blocked with furniture, a small object or a person. These cases have to be handled separately. 
In the Object Perception benchmark (FBM1) the robot has to distinguish between 11 objects in 4 classes and estimate a pose. The pose is evaluated with a motion capturing system to calculate a pose error. Unfortunately this benchmark was skipped due to time constraints.
For the Navigation benchmark (FBM2) the robot has to go to random positions in the apartment that are
given by a refbox. A marker mounted at the robot estimates a ground-truth pose of the robot. When the robot states that it reached the position the ground-truth position is compared against the given goal for translation and rotation separately.



\subsection{Results}



\begin{table}[]
\caption{Results Task Benchmarks}
\centering
\begin{tabular}{|l|r|r|r|r|}
\hline
\textbf{Task}                      & \textbf{A (us)}          & \textbf{P (us)} & \textbf{A (other)} & \textbf{B (other)} \\ \hline
TBM1     & 5                              & 0      & 5 & 2 \\ \hline
TBM2   & 6                              & 0      & 8 & 1 \\ \hline
TBM3  & 4                              & 1      & 2 & 1 \\ \hline
TBM4  & 6                              & 0      & 6 & 0 \\ \hline
\end{tabular}
\label{tab:tbm_results}
\end{table}

\begin{table}[]
\caption{Results Functionality Benchmark FBM2 Navigation}
\centering
\begin{tabular}{|l|r|r|r|}
\hline
\textbf{Task}                      & \textbf{Position (m)}          & \textbf{Orientation (rad)} & \textbf{Hits} \\ \hline
us                  & 0.12 & 0.09 & 0 \\ \hline
other & 3.38 & 2.63 & 1 \\ \hline
\end{tabular}
\label{tab:fbm_results}
\end{table}

\begin{figure*}[t] \centering$
  \vspace{0.06in}
  \begin{array}{cc}
      \includegraphics[height=.28\linewidth]{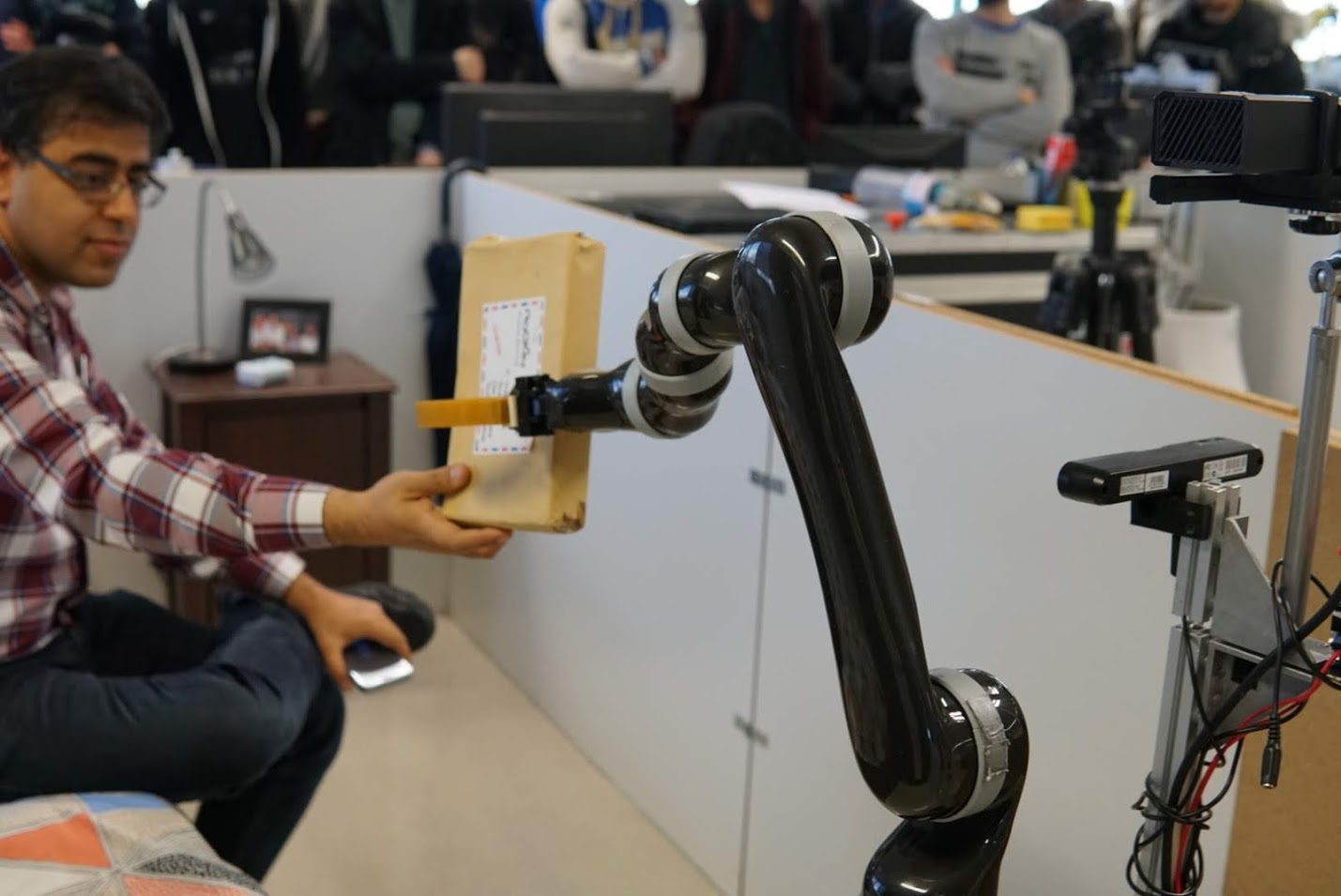} &
      \includegraphics[height=.28\linewidth]{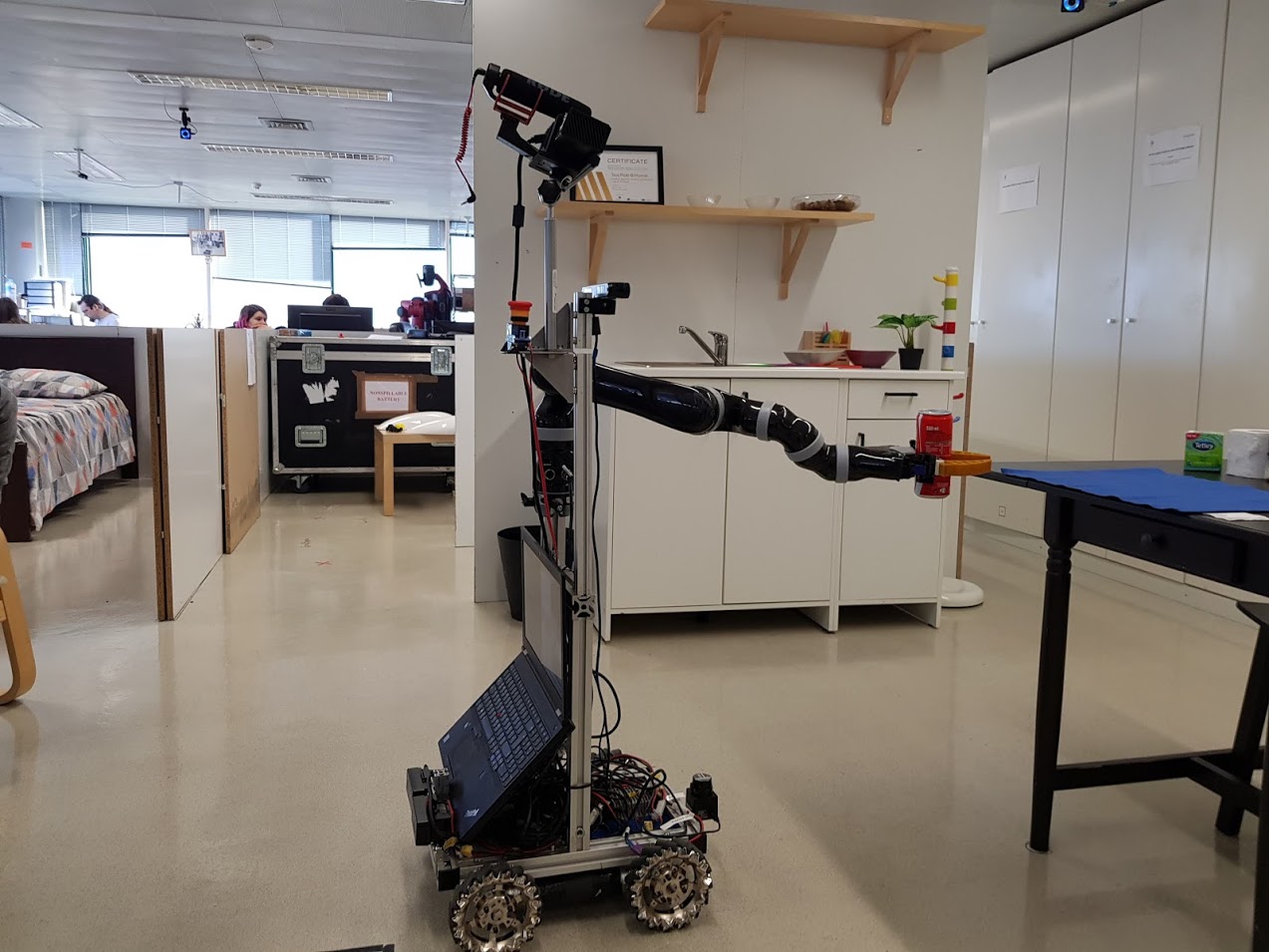} \\
      (a) & (b) 
  \end{array}$
  \caption{Example impressions from \textit{Scratchy} while handing over a parcel to a person during one run of TBM2 and during a test for mobile gripping.}
  \label{fig:runs}
\end{figure*}

Results of the task- and functionality benchmarks are given in \figname \ref{fig:results}. \textit{Scratchy} during a run of TBM2 while handing over a parcel (\figname \ref{fig:runs} $(a)$) and during a test run while gripping (\figname \ref{fig:runs} $(b)$).
We showed during the participation that \textit{Scratchy} is able to compete in the competition environment against already established robots \cite{mateus2015socrob}. Results of our team using \textit{Scratchy} and the competing team are given in Table \ref{tab:tbm_results}. \textit{A} stands for achievements (for completion of a sub-task) and \textit{P} represents the penalties (i.e. for touching an obstacle or wall). For more information about the scoring and the benchmarks we refer to the rulebook \cite{erl_rulebook}. The results represent the mean of the top three scored rounds. The proposed modular robot achieved better results in TBM1 and TBM3. In TBM4 both robots are
on tie but the competing team was able to fulfill the task in a shorter amount of time. 
Most results are improving over the time of the competition (this is especially visible in TBM2 of \figname \ref{fig:results}).
For the functionality benchmark the results are presented in Table  \ref{tab:fbm_results}. We achieved better results in FBM2. However, this table needs to be seen critical as the transformation between the motion capturing marker and the center of the robot were just roughly estimated. Further, the competing team, reported incorrect poses which led to high errors. We still report those numbers here to motivate future organizations to take more care about the benchmarking as we strongly believe in the evaluation concept.

\section{DISCUSSION}

We now want to answer some questions that might come up regarding our proposed robot platform.

In addition to research purposes, we found our approach useful for teaching. The participating students constructed the robot and
designed the schematic descriptions from scratch. Thus they got involved in tasks that are
usually already solved for standard platforms. Through its open source it should be possible to
rebuild and to be made accessible to many people. We provide documentation and rudimentary software 
packages. The transportation costs of the participants and the robot was below 300\euro{} which is mainly credited by the compact disassembled measurements and the low weight.

Is \textit{Scratchy} able to stand a chance against other robots in competitions? We focused on proposing a robot 
that should, despite its minimalist setup, be able to compete in a variety of robot 
competitions. It fits the allowed platform range for RoboCup@Home, RoboCup@Work, as well as European Robotics League for Customer and Professional Service Robots. Even so they may vary in their tasks and
working ranges, \textit{Scratchy} can be adapted because of its flexibility.

Is \textit{Scratchy} usable as a research platform? The flexibility that is provided 
is also beneficial for research. Imaginable are research fields like sensor data fusion,
active semantic mapping, mobile tracking and further. By design we did not have focused
on social robotic fields where the appearance of a robot is important. However, with 
a higher effort \textit{Scratchy} can be also usable in those fields but its not primarily intended.


\section*{ACKNOWLEDGMENT}

We want to thank the students Florian Polster and Malte Roosen who both
contributed to the initial design of the robot base platform, Niklas Yann Wettengel for contributing a lot to the current design and hardware support. Further 
Thomas Weiland and Niko Schmidt for support in the preparation of the TBM4.

\bibliographystyle{IEEEtran}

\bibliography{main}

\end{document}